\documentclass[runningheads]{llncs}

\usepackage{eccv}
\usepackage[width=122mm,left=12mm,paperwidth=146mm,height=193mm,top=12mm,paperheight=217mm]{geometry}
\usepackage{eccvabbrv}

\usepackage{graphicx}
\usepackage{booktabs}

\usepackage[accsupp]{axessibility}  % Improves PDF readability for those with disabilities.

\usepackage{hyperref}

\usepackage{orcidlink}

\begin{document}

\title{DreamMover: Leveraging the Prior of \\ Diffusion Models for Image Interpolation\\ with Large Motion} 

\titlerunning{DreamMover}

% \author{Liao Shen\inst{1}\orcidlink{0000-0002-2423-4835} \and
% Tianqi Liu \inst{1}\orcidlink{0009-0003-0718-0614} \and
% Huiqiang Sun \inst{1}\orcidlink{0000-0002-3653-3613} \and
% Xinyi Ye \inst{1}\orcidlink{0009-0009-1126-3336} \and
% Baopu Li \inst{1}\orcidlink{0000-0002-9032-3991} \and
% Jianming Zhang \inst{2}\orcidlink{0000-0002-9954-6294} \and
% Zhiguo Cao \inst{1}\thanks{Corresponding author}\orcidlink{0000-0002-9223-1863}} 
\author{Liao Shen\inst{1} \and
Tianqi Liu \inst{1} \and
Huiqiang Sun \inst{1} \and
Xinyi Ye \inst{1} \and
Baopu Li \inst{1} \and
\\Jianming Zhang \inst{2} \and
Zhiguo Cao \inst{1}\thanks{Corresponding author}} 
\authorrunning{Liao Shen et al.}

\institute{School of AIA, Huazhong University of Science and Technology\\
\email{\{leoshen,tq\_liu,shq1031,xinyiye,zgcao\}@hust.edu.cn}\\ \email{bpli.cuhk@gmail.com}\\ \and
Adobe Research\\
\email{jianmzha@adobe.com}}

\maketitle
\begin{abstract}
We study the problem of generating intermediate images from image pairs with large motion while maintaining semantic consistency. 
Due to the large motion, the intermediate semantic information may be absent in input images.
Existing methods either limit to small motion or focus on topologically similar objects, leading to artifacts and inconsistency in the interpolation results.
To overcome this challenge, we delve into pre-trained image diffusion models for their capabilities in semantic cognition and representations, ensuring consistent expression of the absent intermediate semantic representations with the input. To this end, we propose \textbf{DreamMover}, a novel image interpolation framework with three main components: 1) A natural flow estimator based on the diffusion model that can implicitly reason about the semantic correspondence between two images. 2) To avoid the loss of detailed information during fusion, our key insight is to fuse information in two parts, high-level space and low-level space.
3) To enhance the consistency between the generated images and input, we propose the self-attention concatenation and replacement approach.
Lastly, we present a challenging benchmark dataset called~\emph{InterpBench} to evaluate the semantic consistency of generated results.
Extensive experiments demonstrate the effectiveness of our method.
Our project is available at \href{https://dreamm0ver.github.io/}{https://dreamm0ver.github.io}.
\end{abstract}

\section{Introduction}
\begin{figure}
  % \begin{center}
    \centering
  \includegraphics[width=0.98\textwidth]{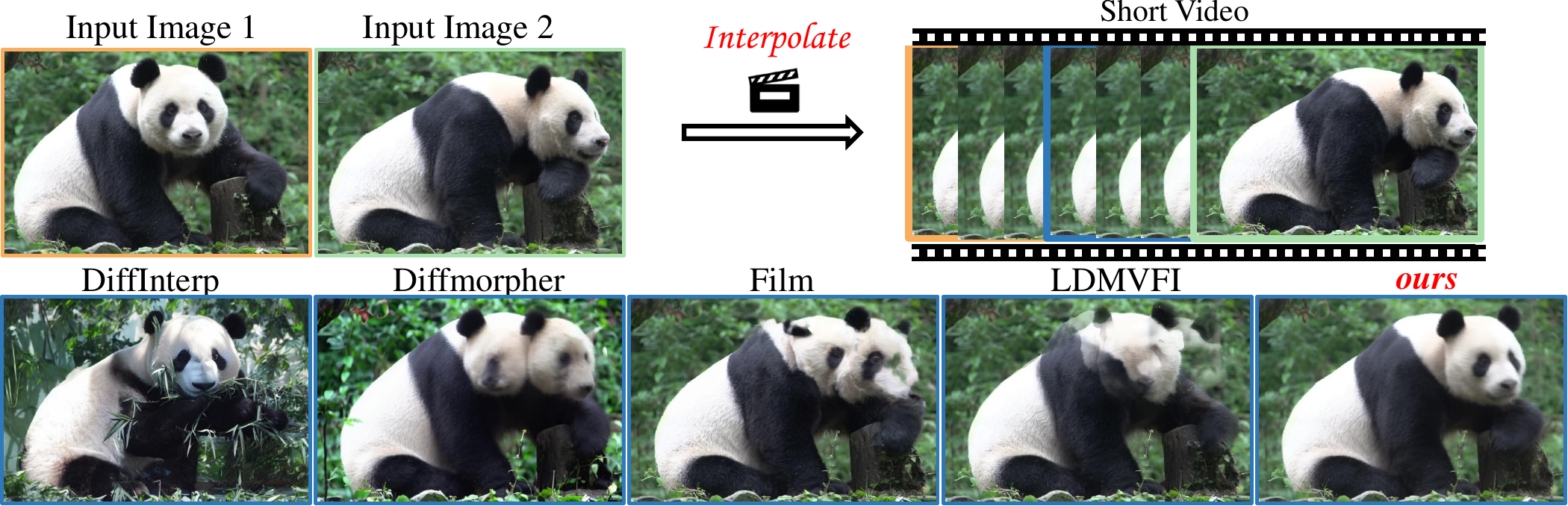}
  \caption{Given two input images with large motion, our proposed method can generate a short video with high fidelity and semantic consistency compared to previous approaches. To see the dynamic effect of our method, we encourage readers to watch our supplementary video.}
  % \end{center}
  \label{fig:teaser}
\end{figure}
With the widespread popularity of short videos on the internet and mobile phone apps such as TikTok and YouTube shorts, people enjoy so much watching short videos. 
The desire for a more engaging visual experience has led to the exploration of innovative technologies in computer vision and graphics, one of which is image interpolation. Image interpolation refers to the process of generating intermediate images from two given images, and it has been a typical and challenging task for many years, especially when these two images show large motions. Two images with large motion captured at different times in one scene often exhibit great variation, and image interpolation aims to recover the potential dynamic processes, providing viewers with lively and dynamic animations. With the input image pair serving as the starting and ending images, such a process generally produces a consistent sequence of object motion videos with rather high fidelity. 

Several existing methods can synthesize intermediate frames from two given images, such as video frame interpolation and image morphing. However, video frame interpolation~\cite{zhang2023extracting,zhewei2020rife,lu2022video} is primarily designed to increase video frame rates, which is significantly different from our purpose of generating short videos. Due to the small differences between adjacent frames, these algorithms often neglect the semantic consistency between input video frames and synthesized intermediate frames. LDMVFI~\cite{danier2023ldmvfi} struggles in large motion and lacks the ability of semantic cognitive. Film~\cite{reda2022film} attempts to interpolate frames between two images with relatively large motion. However, it also operates within near-duplicates and does not model the semantic consistency of intermediate frames. 
On the other hand, image morphing methods~\cite{wang2023interpolating,zhang2023diffmorpher,yang2023impus} can also produce intermediate images from given pairs. However, these models usually focus on the transition between topologically similar objects. 
In contrast, image interpolation mainly aims to construct semantic consistency for the intermediate and input images, generating realistically consistent videos of object movements from two images. The lack of semantic cognitive in the aforementioned methods results in a tendency to split the complete object during interpolation. When applied to such settings, they often result in severe semantic errors and artifacts, leading to inaccuracies in generating intermediate images (as illustrated in Fig.~\ref{fig:teaser} with the erroneous expression of the panda head).

The rise of diffusion models~\cite{ho2020denoising,song2020score,song2020denoising} has made a profound impact on the field of image generation and image editing.
Thanks to the powerful architecture and large aligned image-text datasets~\cite{schuhmann2022laion}, the pre-trained generative diffusion models contain rich implicit semantic information. 
When there is large motion between image pairs, intermediate semantic information may not be present in either of the input images. 
In order to guarantee a coherence transition from one image to the other, we attempt to leverage the pre-trained diffusion model to express the semantic information of input image pairs, and generate intermediate images with high semantic consistency. 

To this end, we propose DreamMover, a novel image interpolation algorithm based on a text-to-image diffusion model, which enables generating large motion videos with semantic consistency from two images. To ensure semantic consistency between the generated and input images, we suggest a new scheme that consists of flow estimation and image fusion.
Specifically, we extract feature maps of the input image pair from the up-blocks of U-Net~\cite{ronneberger2015u} during the noise-adding process. These features are then used to establish pixel correspondences between two images by calculating the cosine distance, further yielding bidirectional optical flow maps. Based on this, we fuse the image pair using softmax splatting~\cite{niklaus2020softmax} and time-weighted interpolation in latent space to generate intermediate images. 

For image fusion, we observed that directly using weighted average operations in latent space may result in a significant loss of high-frequency information, which is not beneficial to modeling semantic consistency. To address this issue, we divide the noisy latent code into two components: a high-level part for overall spatial layout information and a low-level part representing high-frequency details. For the high-level part, we maintain the fusion method using softmax splatting and time-weighted interpolation. For the low-level part, we employ the Winner-Takes-All (WTA) method for fusion. This approach preserves the correct semantic overall layout in the generated video while effectively retaining high-frequency detail information. During the denoising stage, to further ensure semantic consistency, we concatenate the key and value of the input image pairs and replace those of the intermediate ones. Also, we perform low-rank adaptations (LoRAs)~\cite{hu2021lora} to enhance consistency by fine-tuning the diffusion model.

To the best of our knowledge, we are the first image interpolation method considering semantic consistency, which has a vital impact on video effect. Due to the lack of suitable datasets for image interpolation, we curate a dataset, \emph{InterpBench}, to evaluate the performance of generated videos from image interpolation algorithms. Extensive experiments demonstrate that our approach significantly outperforms the state-of-the-art video frame interpolation and image morphing methods. We also conduct a user study to demonstrate the superiority of our method in the view of humans. 

In summary, we propose a novel image interpolation framework that can generate semantic consistent intermediate images from image pairs with large motion, which has the following contributions: 1) a natural optical flow estimator for large motion, 2) a two-level fusion strategy to minimize the loss of high-frequency information, 3) a self-attention concatenation and replacement method to enhance semantic consistency.

% \begin{itemize}[leftmargin=*]
%     \item 
%     \item
%     \item 
% \end{itemize}

\section{Related work}
% \subsection{Image Interpolation}
\textbf{Image Interpolation}.
Previous methods such as video frame interpolation and image morphing can synthesize interpolation images from two given images. Video frame interpolation~\cite{liu2017video,zhang2023extracting,kong2022ifrnet,sim2021xvfi,park2021asymmetric,figueiredo2023frame} are commonly used for up-scale frame rates of videos, which mainly exhibit small motion between consecutive video frames. These methods often lack semantic-level cognitive capabilities and are challenging for large motion.
LDMVFI~\cite{danier2023ldmvfi} trains a diffusion model for video frame interpolation from scratch, but artifacts tend to occur when there is large motion between images.
Film~\cite{reda2022film} attempts to capture relatively large motion in near-duplicate images. However, when the motion is even larger, artifacts and fragmentation often appear. In contrast, our method leverages the prior in pre-trained text-to-image diffusion models and generates reasonable and high-fidelity interpolated images.
DiffInterp~\cite{wang2023interpolating} tries to interpolate images through latent code interpolations and text embedding interpolations. Further, Diffmorpher~\cite{zhang2023diffmorpher} applies low-rank adaptations (LORA)~\cite{hu2021lora} to two images separately and interpolates between the LoRA parameters for semantic transition. However, they mainly focus on two images of topologically similar objects, but may not work well in the same objects with large motion. Unlike them, we use optical flow to fuse information between two images instead of simply overlaying it.

\noindent\textbf{Controllable Image Editing}
Controllable image editing based on diffusion model is a challenging task that aims to manipulate and generate novel images according to various conditions, including text-based editing~\cite{hertz2022prompt,tumanyan2023plug,epstein2023diffusion}, image-based editing~\cite{saharia2022palette,zhang2023adding,meng2021sdedit,shen2023make}, point-based editing~\cite{pan2023drag,shi2023dragdiffusion,mou2023dragondiffusion} and motion-based editing~\cite{geng2024motion,shi2024motion}. 
These methods mainly add noise to the clean image using DDIM inversion~\cite{song2020denoising} and denoise by the guidance of various conditions. In this way, diffusion models can generate high-quality new images that fit well with the semantics of the origin image. 
Most of these works edit a single image and generate semantically consistent edited images, while the generation of intermediate results from two images is much less explored in image diffusion models. 

\noindent\textbf{Image-to-Video Diffusion Models}
Previous works on Image-to-Video Diffusion Models~\cite{singer2022make,zeng2023make,xing2023dynamicrafter,yu2023animatezero} have achieved great success, which contain downstream tasks that can be used for frame interpolation between two images. We differ significantly from these methods in that we edit images to generate intermediate image sequences via the prior pre-trained image diffusion models, but they directly utilize video diffusion models which require more complex architectures and training on large-scale video datasets.
\section{Method}
\begin{figure}
    \centering
    \includegraphics[width=1.0\linewidth]{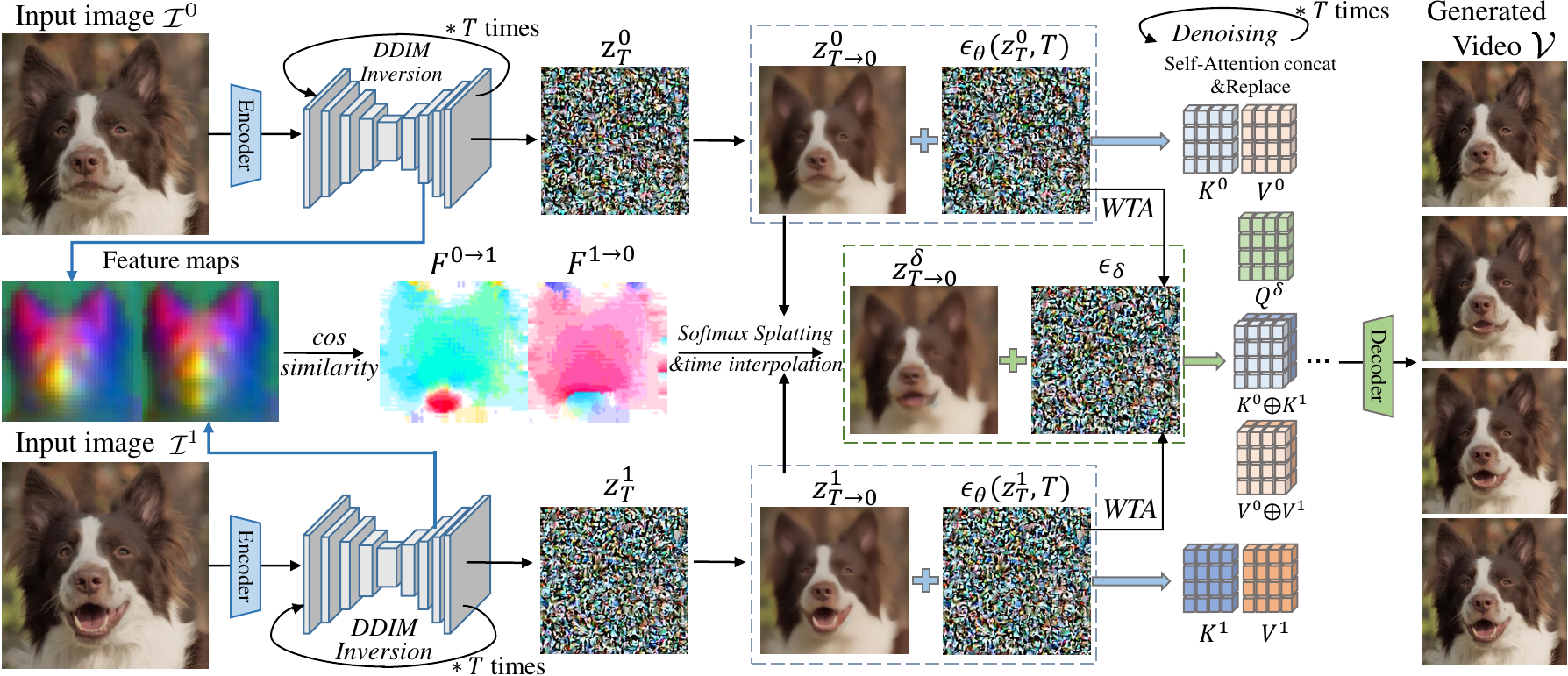}
   \caption{\textbf{Overview of our method.} Given two input images $\mathcal{I}^0$ and $\mathcal{I}^1$, we extract feature maps and leverage them to obtain the bidirectional optical flow $F^{0\to1}$ and $F^{1\to0}$. Next, we decompose the noisy latent code $z_T$ into two-level space and perform softmax splatting and time interpolation for image fusion. For high-frequency information $\epsilon_\theta$, we replace all weighted average operations with "Winner-Takes-All"(WTA). In addition, we propose a novel self-attention replacement method for consistency. Finally, our method can generate a sequence of high-fidelity interpolation frames.}
\label{fig:pipeline}
\end{figure}

Given a pair of images $\mathcal{I}^0$ and $\mathcal{I}^1$ with large motion, we aim to generate intermediate images $\mathcal{I}^\delta$ and yield a semantically consistent video $\mathcal{V}=\{\mathcal{I}^\delta|\delta \in (0,1)\}$, where the sequence length of time $\delta$ depends on the desired number of interpolation images $n$.

We schematically illustrate our pipeline in Fig.~\ref{fig:pipeline}. Our method starts by obtaining bidirectional optical flow from correspondence between the feature maps (Sec.~\ref{sec:flow}). In order to preserve the details of interpolation images carefully during fusion, we divide the origin latent space into two parts, high-level and low-level space, and operate on each part individually (Sec.~\ref{sec:fusion}). Finally, to enhance the appearance consistency between the two input images, we propose the self-attention concatenation and replacement during denoising, and perform LoRA for semantic-preserving (Sec.~\ref{sec:consistent}). 
% To minimise the loss of high-frequency information, we avoid to directly apply Softmax Splatting and temporal interpolation in latent code space
\subsection{Premininaries}
\noindent\textbf{Latent diffusion model} (LDM)~\cite{rombach2022high} stands out as an efficient variant of diffusion models, employing the diffusion process within the latent space. This involves the implementation of both a forward and a backward process. For a given clean latent input $z_0$, the forward diffusion process gradually adds Gaussian noise at each timestamp $t$ to obtain $z_t$:
$$
q(z_t|z_{t-1}) = \mathcal{N}(z_t;\sqrt{1-\beta_t}z_{t-1},\beta_tI),
$$
where $\{\beta_t\}^T_{t=1}$ represent the scale of noises, and $T$ denotes the number of diffusion timestamps. Then the backward denoising process utilizes a trained U-Net $\epsilon_\theta$ for denoising:
$$
p_\theta(z_{t-1}|z_t) = \mathcal{N}(z_{t-1};\mu_\theta(z_t,t),\Sigma_\theta(z_t,t)),
$$
where $\mu_\theta$ and $\Sigma_\theta$ are computed by $\epsilon_\theta$.

To accurately reconstruct given real images, we employ the deterministic DDIM inversion and sampling~\cite{song2020denoising} to add noise and remove noise.
We can simplify the denoising process into the following form to predict the $z_{t-1}$ of previous timestamp:
\begin{equation}
z_{t-1} = \sqrt{\alpha_{t-1}} \cdot {z_{t \to 0}} +\sqrt{1-\alpha_{t-1}} \cdot \epsilon_\theta(z_t,t),  \tag{2} \label{eq2}
\end{equation}
\begin{equation}
{z_{t \to 0}} = \frac{z_t-\sqrt{1-\alpha_t}\epsilon_\theta(z_t,t)}{\sqrt{\alpha_t}}. \tag{3} \label{eq3}
\end{equation}
where $t$ denotes the noisy time, $\alpha_t=\prod_{i=1}^t(1-\beta_i)$  and $z_{t \to 0}$ means the predicted clean latent code that is directly denoised from $z_t$.
\begin{figure}[!t]
  \centering
  \setlength{\belowcaptionskip}{-5pt}
  \includegraphics[width=0.9\textwidth,height=4cm]{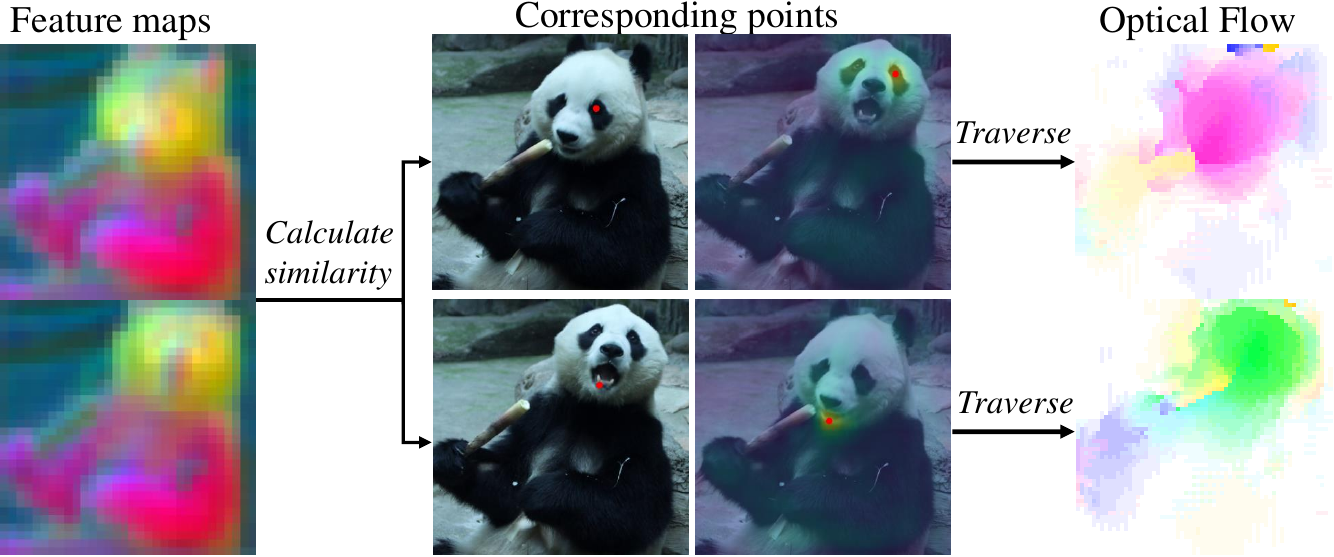}
  \caption{\textbf{The potential of diffusion model for optical flow estimation.} We perform PCA on the features and observe consistent spatial layouts with input images, and obtain bidirectional optical flow through the correspondence between feature maps.}
  \label{fig:flow}
\end{figure}
\subsection{Diffusion-aware flow estimation}
\label{sec:flow}
Given two images $\mathcal{I}^0$ and $\mathcal{I}^1$, optical flow estimation is a key step in image interpolation, which indicates the correspondences of pixels between two images and can be employed to warp pixels to generate the intermediate results.
We can warp an image with an optical flow $F$ by softmax splatting method~\cite{niklaus2020softmax}:
\begin{equation}
    \overrightarrow{\sigma}(\mathcal{I},F) = \frac{\sum(exp(M) \cdot \mathcal{I}, F)}{\sum (exp(M),F)},  \tag{4} \label{eq4}
\end{equation}
where $M$ is a metric of brightness constancy~\cite{baker2011database}.

\begin{figure}[!t]
  \centering
  \setlength{\belowcaptionskip}{-5pt}
  \includegraphics[width=\linewidth]{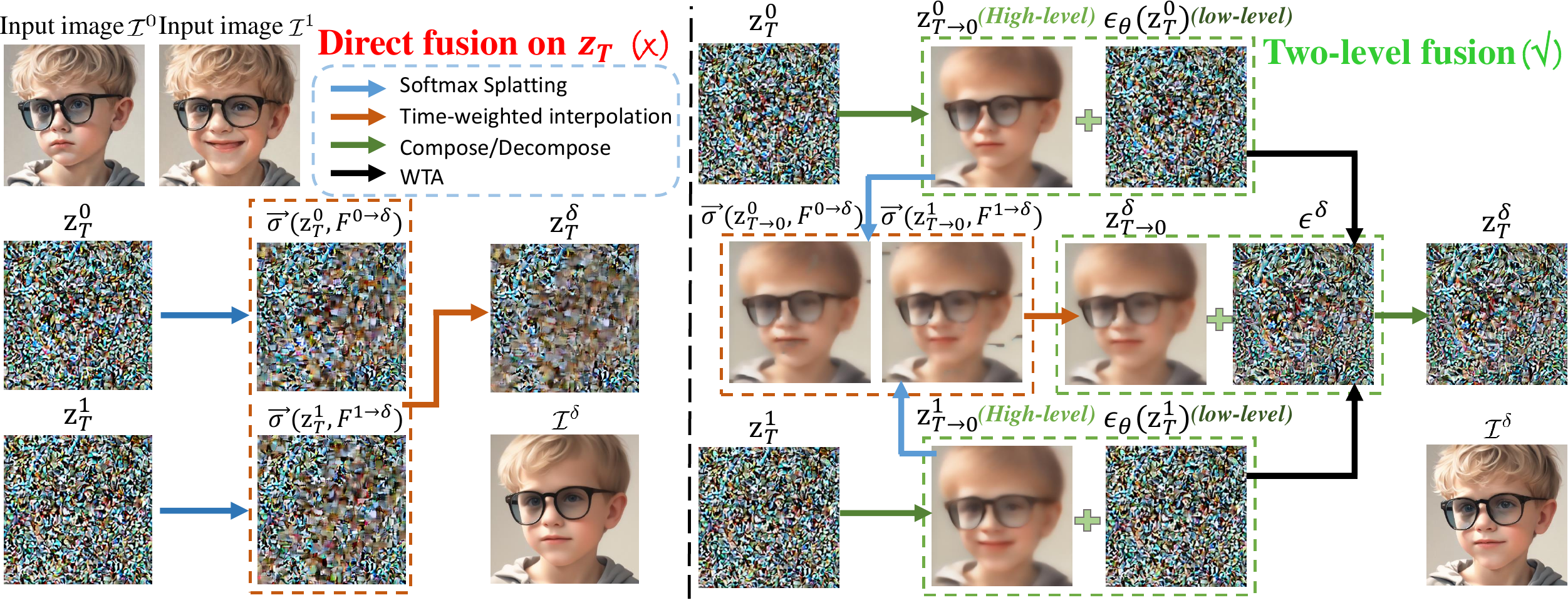}
  \caption{\textbf{The process of direct fusion and our proposed two-level fusion.} Generally, $z_{T\to0}$ represents a latent code. Here, for clearer visualization, we illustrate the RGB image decoded from it to emphasize a significant loss of high-frequency information compared to input images.
  % Normally, $z_{T\to0}$ is a latent code, and we visualise its RGB image in the figure in order to show that it has only a small amount of high-frequency information compared to input images.
  }
  \label{fig:fusion1}
\end{figure}
Specifically, we encode $\mathcal{I}_0$ and $\mathcal{I}_1$ into the latent space to get $z^0$ and $z^1$. 
By getting a bidirectional optical flow $F^{0\to1}$ and $F^{1\to0}$ from the two images, we can warp $z^0$ and $z^1$ to the middle time $\delta \in (0,1)$ using softmax splatting, and get the middle latent code $z^{0\to \delta}$ and $z^{1\to \delta}$ respectively. The final intermediate latent code $z^\delta$ can be obtained by fusing them with time-weighted interpolation.

The crux of the matter lies in getting bidirectional optical flow between two images without introducing additional optical flow prediction modules. 
Drawing inspiration from~\cite{tang2023emergent,luo2023diffusion,zhang2023tale}, diffusion model engages in implicit reasoning about image correspondences, yielding remarkably robust and accurate results. Therefore, we use the pre-trained diffusion model to obtain optical flow through semantic correspondence between real images, without the need for additional fine-tuning or supervision.
Specifically, we employ the DDIM inversion~\cite{song2020denoising} to send $z^{0}_0$ and $z^{1}_0$ into the U-Net for adding noise, where $z^{\delta}_t$ represents the latent code of intermediate time $\delta$ after $t$ steps of noise-adding. Meanwhile, feature maps $f^0$ and $f^1$ are extracted from up-blocks of U-Net.
% Specifically, we add noise to $z^{0}_0$ and $z^{1}_0$ using DDIM inversion ~\cite{song2020denoising} before passing it into the U-Net to extract feature maps $f^0$ and $f^1$ from up-blocks. As for $z^{\delta}_t$, $\delta$ means interpolation time and $t$ means noisy timestamp.
As illustrated in Fig.~\ref{fig:flow}, the spatial layout between the feature maps is highly similar to that of the original images, which provides us with the possibility for optical flow prediction of two images in latent space. 
By traversing the points in one feature map, we can select the pixel in the other map with the highest cosine similarity as its corresponding location, thereby obtaining the bidirectional optical flow maps: 
% In this way, we can acquire the flow from one image to the other on latent space.
\begin{equation}
F^{0 \to 1}(x,y) = \mathop{\arg\max}\limits_{i,j}\langle f^0(x,y), f^1(i,j)\rangle , \tag{5} \label{eq5}
\end{equation}
where $(x,y)$ and $(i,j)$ are the indexes of $I^0$ and $I^1$, and $\langle \  ,\ \rangle$ denotes the calculating of cosine similarity. 
Likewise, we can derive the optical flow $F^{1 \to 0}$ from $I^1$ to $I^0$.
Lastly, we can gain the optical flow $F^{0\to \delta} = \delta \cdot F^{0\to 1}$ from time $0$ to the intermediate time $\delta$ and the flow $F^{1\to \delta} = (1-\delta) \cdot F^{1\to 0}$ from time $1$ to the intermediate time $\delta$.

\begin{figure}[!t]
  \centering
  \setlength{\belowcaptionskip}{-5pt}
  \includegraphics[width=\linewidth]{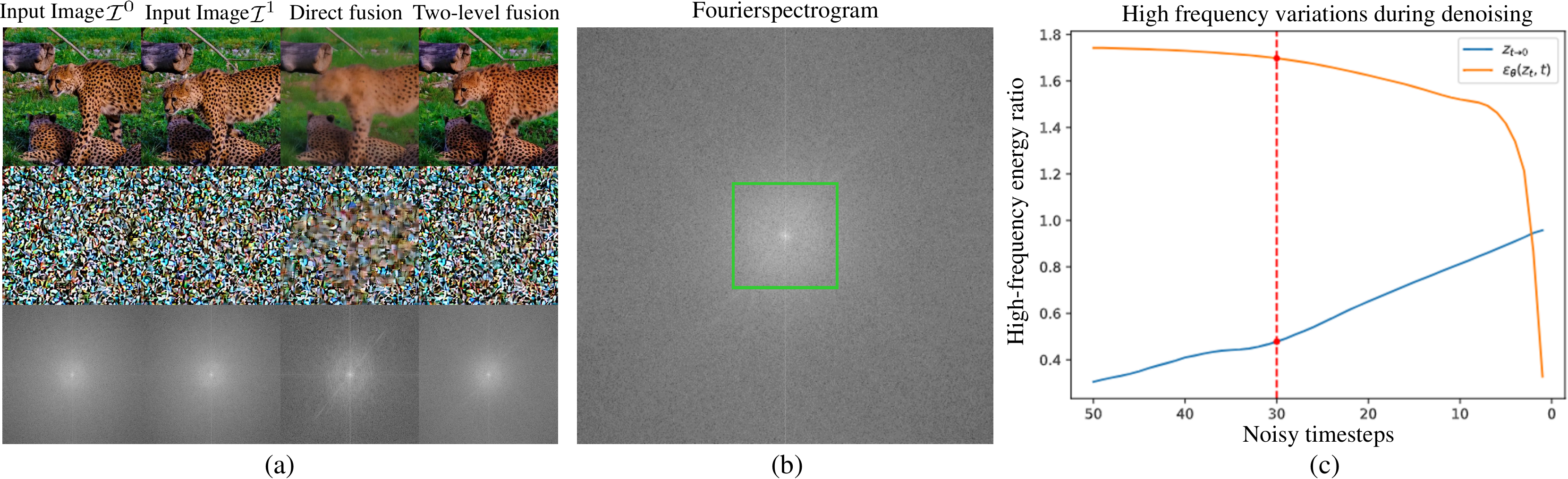}
  \caption{\textbf{(a) Effects of fusion in different space.} Compared to direct fusion, our strategy better preserves details in the RGB image and maintains more high-frequency energy in the Fourier spectrograms. \textbf{(b) Definition of high-frequency region.} We define it as the part of the spectrogram beyond the centre 1/4. \textbf{(c) High-frequency variations during denoising.}
  }
  \label{fig:fusion}
\end{figure}
\subsection{Latent space fusion}
\label{sec:fusion}
In order to combine the effective semantic information of the input image pairs, we apply softmax splatting~\cite{niklaus2020softmax} to respectively warp these two images to middle time $\delta$ based on the bi-directional flow, and acquire the middle results $z^{0\to\delta}=\overrightarrow{\sigma}(z^0, F^{0 \to \delta})$ and $z^{1\to\delta}=\overrightarrow{\sigma}(z^1, F^{1 \to \delta})$. Further, we can fuse them together using time-weighted interpolation. 
However, due to the large object motion between two input images, the intermediate semantic information may be absent in input images. Hence, we leverage the potential semantic capability of image diffusion to yield reasonable results and maintain semantic consistency.
% Typically, we can apply Softmax Splatting to respectively warp two images to middle time $\delta$ based on the bi-directional flow, and acquire the middle results $z^{0\to\delta}=\overrightarrow{\sigma}(z^0, F^{0 \to \delta})$ and $z^{1\to\delta}=\overrightarrow{\sigma}(z^1, F^{1 \to \delta})$. They then can be transmitted to a refinement network, and the ultimate result $z^\delta$ is synthesized. However, it is worth noting that we explore the potential semantic capability of pre-trained diffusion to realise the fusion process and yield the final interpolation images without training a new module.

\noindent\textbf{Direct fusion.}
We first add noise through DDIM inversion to $z^{0}_0$ and $z^{1}_0$ in order to obtain $z^{0}_T$ and $z^{1}_T$, where $T$ is the total number of noise addition steps. Then, we simply warp them to the middle and combine them according to time-weighted interpolation to obtain the noisy latent code for the middle time $\delta$: 
\begin{equation}
z^{\delta}_T = (1-\delta) \cdot \overrightarrow{\sigma}(z^{0}_T, F^{0\to \delta}) + \delta \cdot \overrightarrow{\sigma}(z^{1}_T, F^{1\to \delta}). \tag{6} \label{eq6}
\end{equation}
Finally, we transmit $z^{\delta}_T$ to the diffusion model for denoising and desire to directly generate a reasonable and fidelity result.
Nevertheless, as demonstrated in Fig.~\ref{fig:fusion1}, the results of direct fusion exhibit noticeable blurriness. Intuitively, both softmax splatting and interpolation will introduce average operations, leading to the loss of high-frequency information. We provide mathematical explanations and conduct a thorough analysis in supplementary material.

\noindent\textbf{Two-level fusion.} 
Revisiting the denoising process of diffusion as shown in Eq.~\ref{eq2}, it has two components $z_{t \to 0}$ and $\epsilon_\theta(z_t,t)$.
% , where $z_{t \to 0}$ is a predicted clean latent directly from noisy latent code $z_t$. 
% Since it is a one-step denoising rather than a multi-step progressive denoising, $z_{t \to 0}$ lacks high-frequency information compared to clean latent $z_0$ as illustrated in Fig.~\ref{fig:fusion1}. 
For $z_{t \to 0}$, since it is a predicted clean latent by a one-step denoising rather than a multi-step progressive denoising, it can only capture certain high-level context information while lacking high-frequency details. On the other hand, the component $\epsilon_\theta(z_t,t)$ serves to complement low-level textures during denoising. 
we show quantitatively that $\epsilon_\theta(z_t,t)$ has more high-frequency components than $z_{t \to 0}$ and that our strategy retains more high-frequency information in Fig.~\ref{fig:fusion}(c).We define high-frequency energy by the sum of the amplitudes in the high-frequency region of the Fourier spectrogram. For the high-frequency region, we define it as the part of the spectrogram beyond the centre 1/4. Specifically, the portion outside the green square in Fig.~\ref{fig:fusion}(b). We take the ratio of their respective high-frequency energies to that of input image as a metric(y-axis). We perform fusion operations at the $30$th noisy step, and $\epsilon_\theta(z_t,t)$ has more high-frequency information compared to $z_{t\to0}$.

To integrate the effective information of two images while preserving high-frequency details, we propose a two-level fusion strategy.
Specifically, for high-level information, we perform fusion in the $z_{T\to 0}$ space.
% Further, our key insight is that $z_{t \to 0}$ provides high-level textures like spatial layout while $\epsilon_\theta(z_t,t)$ complements the low-level textures during denoising. We therefore abandon the direct fusion on $z^{\delta}_{T}$ space and instead perform fusion on $z^{\delta}_{T\to 0}$. In this case, we can minimise the loss of high-frequency information. 
Per Eq.~\ref{eq3}, $z^{0}_{T\to0}$ and $z^{1}_{T\to0}$ can be obtained from $z^{0}_T$ and $z^{1}_T$, and the fused result is:
\begin{equation}
z^{\delta}_{T\to0} = (1-\delta) \cdot \overrightarrow{\sigma}(z^{0}_{T\to0}, F^{0\to \delta}) + \delta \cdot \overrightarrow{\sigma}(z^{1}_{T\to0}, F^{1\to \delta}). \tag{7} \label{eq7}
\end{equation}
For low-level information, we perform fusion on $\epsilon_\theta$ space. To mitigate the loss of high-frequency information caused by average operation in softmax splatting and time interpolation, we apply the "Winner-Takes-All"(WTA) operation, i.e, taking values of the highest weights, to obtain the fused result $\epsilon^\delta = WTA(\epsilon_\theta(z^{0}_T),\epsilon_\theta(z^{1}_T))$.
After obtaining the separately fused results for two levels, we backtrack to obtain $z^{\delta}_{T}$,
 \begin{equation}
z^{\delta}_{T} = \sqrt{\alpha_T} \cdot z^{\delta}_{T\to0} + \sqrt{1-\alpha_T} \cdot \epsilon^\delta. \tag{8} \label{eq8}
\end{equation}
% Median filtering is also used to eliminate isolated points.
% At this point, we get the predicted clean latent $z^{\delta}_{T\to0}$ at the time $\delta$, and then based on Eq.~\ref{eq3}, we can backtrack to obtain $z^{\delta}_{T}$,
%  \begin{equation}
% z^{\delta}_{T} = \sqrt{\alpha_T} \cdot z^{\delta}_{T\to0} + \sqrt{1-\alpha_T} \cdot \epsilon^\delta, \tag{8} \label{eq8}
% \end{equation}
% where $\epsilon^\delta$ is obtained from $\epsilon_\theta(z^{0}_T)$ and $\epsilon_\theta(z^{1}_T)$. The fusion process for low-level space is similar to the above, but change all the weighted averaging in Softmax Splatting and time interpolation to "Winner-Takes-All"(WTA) operations, i.e, take the value of whoever has the highest weights, in order to avoid losses bringing from averaging operations. Median filtering is also used to eliminate isolated points.
Finally, based on Eq.~\ref{eq2}, we can yield a clean latent $z^{\delta}_0$ by performing $T$ times of denoising, and then send it to the decoder of diffusion to get the intermediate image $\mathcal{I}^\delta$. 
As shown in Fig.~\ref{fig:fusion}(a), the quality of images obtained through our proposed two-level fusion is significantly superior to the one of direct fusion.
\subsection{Reference-guided consistency }
\label{sec:consistent}
\begin{figure}[!t]
  \centering
  \setlength{\belowcaptionskip}{-5pt}
  \includegraphics[width=0.9\linewidth]{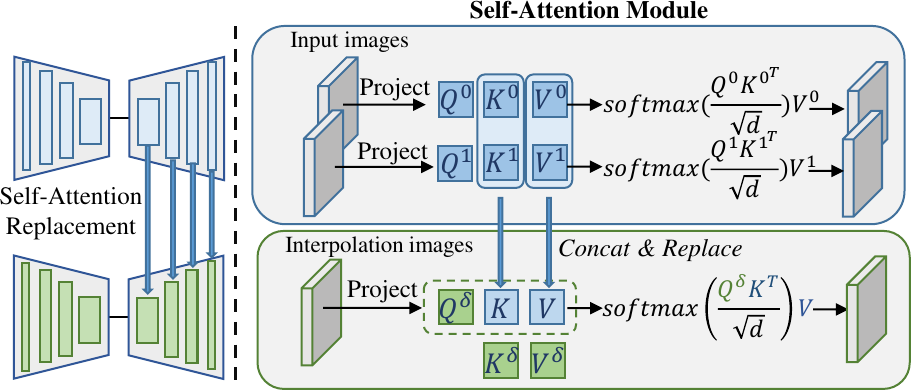}
  \caption{\textbf{Self-attention concatenation and replacement.} }
  \label{fig:attention_control}
\end{figure}
Although the intermediate results are reasonable regarding spatial layout, we observe inconsistent changes in the generated images. We posit that this issue arises due to the absence of adequate guidance from the original input images during the denoising process. To solve this problem, we draw inspiration from attention control techniques in previous image editing research~\cite{cao2023masactrl,parmar2023zero,tumanyan2023plug,chefer2023attend,kawar2023imagic,khachatryan2023text2video} and propose a novel self-attention concatenation and replacement method, which introduces the attention features of the input image pair during denosing into the denoising process of the intermediate image. Specifically, we can use the query features in the self-attention module of interpolation images to query the corresponding key and value features in input image pairs.
% Thus, intermediate latent codes can query correlated local textures and structures from input image pairs to further boost semantic consistency.

As shown in Fig.~\ref{fig:attention_control}, in the denoising steps, we feed the noisy latent code of the input two images into the U-Net to obtain the key and value matrices $K^i, V^i(i = 0, 1)$ in the self-attention modules of U-Net up-sampling blocks. In order to generate a reliable intermediate image $I_\delta$, we replace its key and value by concatenating $K^i$ and $V^i$:
\begin{equation}
Q = Q^\delta,\quad K = (K^0 \oplus K^1),\quad V = (V^0 \oplus V^1) ; \tag{9} \label{eq9}
\end{equation}
\begin{equation}
Attention^\delta = softmax(\frac{QK^T}{\sqrt{d_k}})V, \tag{10} \label{eq10}
\end{equation}
where $\oplus$ denotes the concatenation operation. Thus, intermediate latent code can query correlated local structures and textures from both input images to further enhance consistency.

%lora这一段可以再精简（sla）
In addition, we conduct Low-Rank Adaption (LoRA)~\cite{hu2021lora} to further improve the semantic consistency of the intermediate images with input images.
% Instead of directly tuning the entire diffusion model, LoRA fine-tunes the model parameters $\theta$ by training a low-rank residual part $\Delta \theta$, where $\Delta \theta$ can be decomposed into low-rank matrices.
% Besides its inherent advantage in tuning efficiency, LoRA enjoys an impressive capacity to encapsulate identity of given images into the low-rank parameter space. 
Unlike Diffmorpher~\cite{zhang2023diffmorpher}, which requires adapting LoRAs to the input two images respectively, our method simply fits a single LoRA for the image pair. Finally, the fine-tuned model can generate samples with consistent semantic identity.

\section{Experiments}
\subsection{Implementation Details}
In all of our experiments, we adopt the Stable Diffusion $1.5$~\cite{rombach2022high} as our diffusion model and the number of interpolation images is $32$. During the latent optimization stage, we schedule $50$ steps for DDIM and optimize the diffusion latent at the $30$th noisy step unless specified otherwise, and we extract the output of the second up-block of the UNet at the $14$th noisy step as feature maps used for flow estimation. In addition, we set the rank of LoRA to $16$. We fine-tune the LoRA using the AdamW optimizer with a learning rate of $5 \times 10^{-4}$ for $80$ steps, and it takes $\sim 40$ seconds on a single NVIDIA RTX $3090$ GPU.
It is noteworthy that in both DDIM inversion and denoising, we do not apply classifier-free guidance (CFG)~\cite{ho2022classifier}. This is because CFG tends to accumulate numerical errors and cause supersaturation problems~\cite{mokady2023null}. 
\subsection{Baselines and Evaluation metrics}
To evaluate the effectiveness of our method, we extensively compare our outcomes with two image morphing techniques and two video interpolation methods. Diffinterp~\cite{wang2023interpolating} and Diffmorpher~\cite{zhang2023diffmorpher} are diffusion model-based methods that can generate a sequence of intermediary images for two given images of topologically similar objects. Film~\cite{reda2022film} trained on multi-scale video interpolation datasets attempts to handle frame interpolation for large motion. LDMVFI~\cite{danier2023ldmvfi} trains a latent diffusion model for video interpolation, since this method just obtains one intermediate image, we iterate to generate a sequence of interpolated images.

To quantitatively evaluate the quality of interpolation images and the generated videos, we adopt Fréchet Inception Distance (FID)~\cite{heusel2017gans}, Perceptual Similarity (LPIPS)~\cite{zhang2018unreasonable}, Warping Error (WE)~\cite{lai2018learning}, and $\text{WE}_{mid}$ as our metrics. We use FID and LPIPS to evaluate the fidelity and rationality of all methods, and utilize WE to evaluate the temporal coherency of the generated videos. In addition, we employ $\text{WE}_{mid}$ to measure whether the middle-most image is consistent with the input image pair.
\subsection{InterpBench}
Due to the lack of discussion on semantic consistency modeling for image interpolation, there are currently no datasets suitable for our task. Existing video frame interpolation datasets~\cite{xue2019video, baker2011database, liu2017video,montgomery2021xiph} provide triplets of input image pairs and intermediate images, but they are designed for scenarios where the motion between two input images is minimal. As such, discussing the semantic consistency of the intermediate image may be meaningless and not applicable to evaluating the performance of image interpolation algorithms. To meet the demand for performance evaluation of image interpolation algorithms, we introduce \emph{InterpBench}, the first benchmark dataset tailored for image interpolation. 
% Since traditional interpolation methods are typically used to boost video frame rates, the previous datasets, mainly had slight motion between frames. 
% Hence, there is an absence of dedicated evaluation benchmarks for the task of large motion interpolation, making it challenging to comprehensively study the effectiveness of our proposed approach. To evaluate the semantic consistency of generated videos , we introduce \emph{InterpBench}, the first benchmark dataset tailored for image interpolation with large in-between motion. 
\emph{InterpBench} is a diverse compilation encompassing various large motions of objects and we collected 100 pairs of pictures in total. Details of our dataset can be found in the supplementary materials.
\begin{figure*}[t]
\begin{center}
    \includegraphics[width=1.0\linewidth]{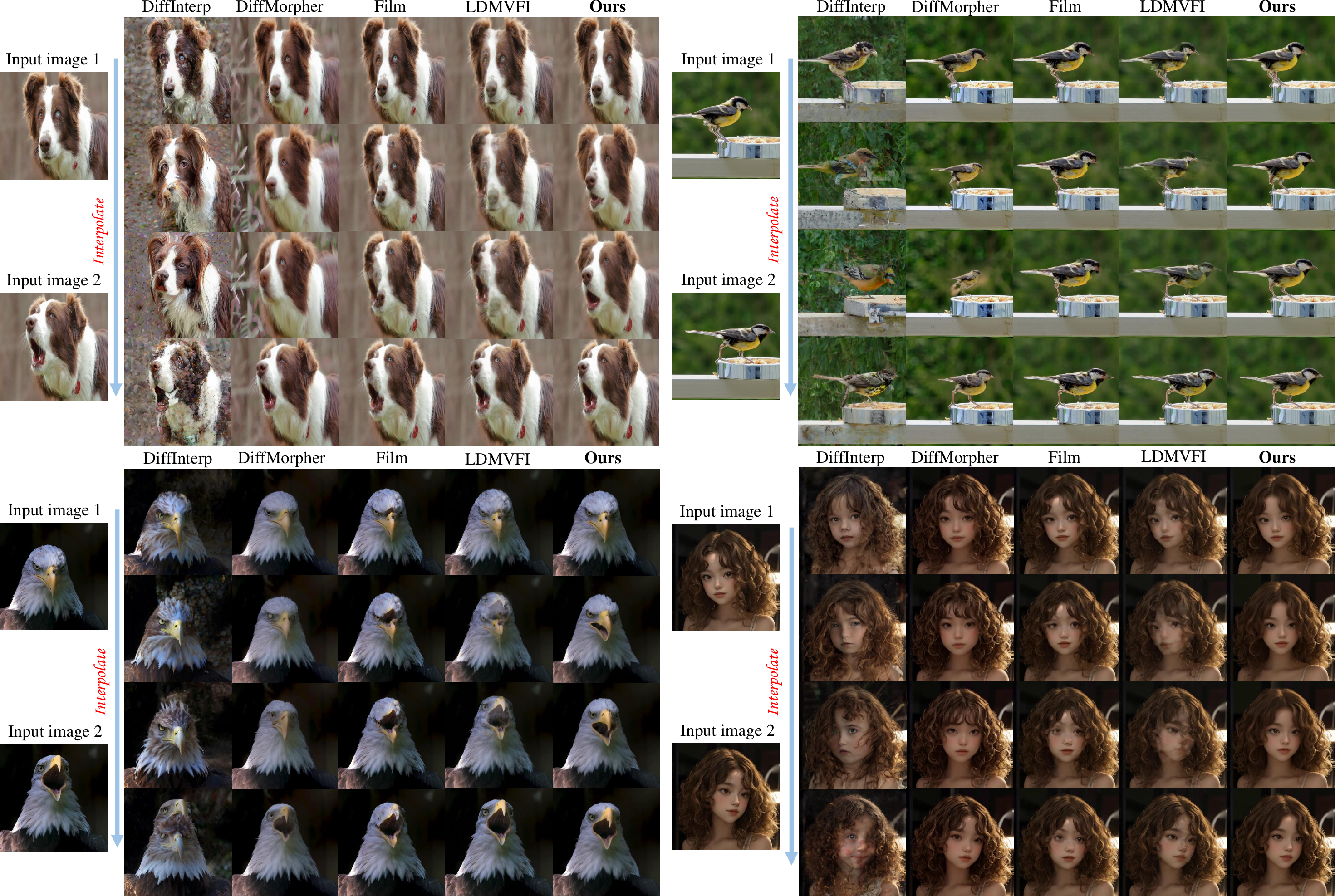}
\end{center}
   \caption{\textbf{Qualitative comparisons of baselines and our method on \emph{InterpBench}.} For each scenario, from left to right we show four methods: Diffinterp, DiffMorpher, Film, and ours, and from top to bottom we show four images interpolated from each method.}
\label{fig:qualitative}
\end{figure*}
% as fidelity and rationality metrics. In addition, we use and Warping error (WE)~\cite{lai2018learning} multiplied by $100$ to evaluate the consistency of middle images. 
% Since most of the large motion in real world is not uniform rigid motion, it is difficult to align the generated interpolation frames with ground-truth intermediate frames, thus making it difficult to calculate PSNR and SSIM metrics. More details are given in supplementary materials.
\begin{figure}[!t]
  \centering
  \setlength{\belowcaptionskip}{-5pt}
  \includegraphics[width=0.98\linewidth]{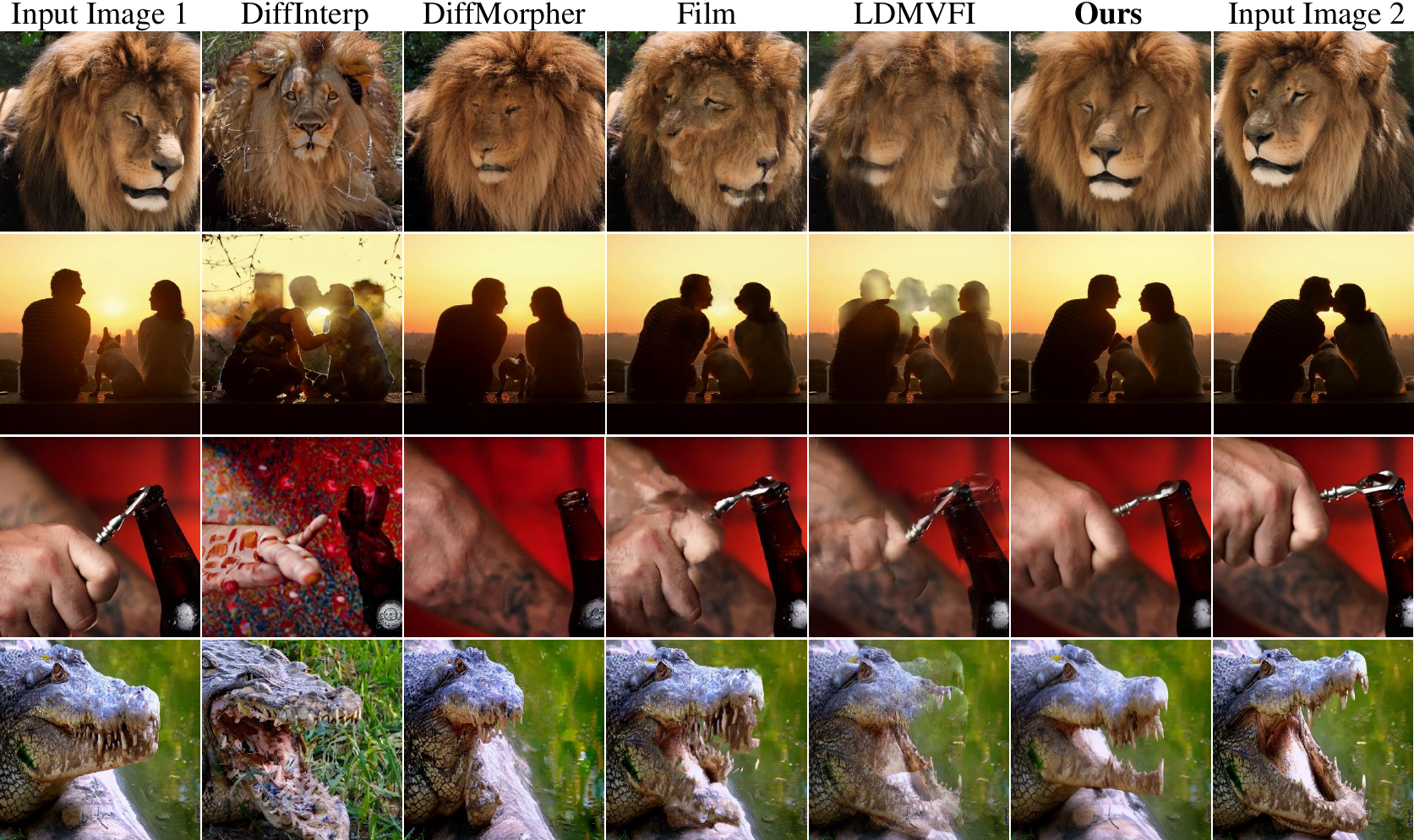}
  \caption{\textbf{More Visualization Comparison of baselines and our method.} We show the middle-most image obtained by all methods. Our approach generates intermediate results that maintain the best semantic consistency.}
  \label{fig:visualisation1}
\end{figure}
\subsection{Qualitative Evaluation}
Visual qualitative comparisons are shown in Fig.~\ref{fig:qualitative}. Our method outperforms all other baselines in terms of image fidelity, image detail, and semantic consistency of interpolated images. In particular, our method can generate reasonable and realistic intermediate results, such as a puppy and an eagle slowly opening their mouths. However, Diffinterp cannot produce results that are consistent with the input and the results are full of flickering artifacts. Diffmorpher cannot handle correct semantic transitions with large object motion, resulting in low-quality and distorted images. The results of Film and LDMVFI produce artifacts and give the impression of fragmentation. Also note the legs of the bird in the top right of Fig.~\ref{fig:qualitative}, only our method retains the details. Meanwhile, we can observe the girl in the bottom right, our results have the best quality and consistency. 

For more comparison results, please see Fig.~\ref{fig:visualisation1} and the supplementary material. We show a variety of scenarios to demonstrate the superiority of our approach in both image details and semantic consistency. Furthermore, we hope readers to watch the supplementary video for a better dynamic comparison.
\subsection{Quantitative Evaluation}
As shown in Table ~\ref{tab:Quantitative comparisons}, our method outperforms all baselines across most metrics by a large margin. 
Specifically, our approach produces higher-quality images with fewer artifacts, resulting in significantly better FID than other approaches.
Additionally, thanks to our effective modeling of semantic representation and consistency, the images generated by our method exhibit higher consistency with input images, achieving optimal LPIPS and $\text{WE}_{mid}$ metrics, which measure the consistency of high-level semantic and low-level detail information, respectively. 
Film~\cite{reda2022film} and LDMVFI~\cite{danier2023ldmvfi} achieve better temporal consistency metrics, but the content they generate does not guarantee semantically consistent representations of objects, which can seriously affect the quality of the generated video. We encourage readers to watch the video of the supplementary material, which can reflect the superiority of our method more intuitively.
\begin{table}[t]
\centering
\begin{minipage}[t]{0.55\linewidth}
\centering
\caption{\textbf{Quantitative comparisons against all baselines on \emph{InterpBench}}. The better approach favors lower FID, LPIPS and WE metrics. The best performance is in \textbf{bold}.}
\label{tab:Quantitative comparisons}
\begin{tabular}{lcccc}
    \toprule    Method&FID$\downarrow$&LPIPS$\downarrow$ &WE $\downarrow$&$\text{WE}_{mid}$ $\downarrow$\\
    \midrule
    DiffInterp   &185.7836 &0.5375  &0.5112 &0.9573\\
    Diffmorpher  &68.2286 &0.3061 &0.2673 &0.7784\\
    Film  &54.2792 &0.2313 &\textbf{0.1244} &0.4176\\
    LDMVFI &48.3469 &0.2347 &0.1453 &0.4373\\
    Ours &\textbf{43.1798} &\textbf{0.2227} &0.2069 &\textbf{0.3687}\\
  \bottomrule
\end{tabular}
\end{minipage}~
\begin{minipage}[t]{0.46\linewidth}
\centering
\caption{\textbf{User study.} Pairwise comparison results indicate that users prefer our method as better quality and fidelity.}
\label{tab:user study}
\begin{tabular}{lc}
    \toprule
    Comparison&Human preference\\
    \midrule
    Diffinterp / \textbf{Ours}  &5.61\% /\textbf{94.39\%}\\
    Diffmorpher / \textbf{Ours} &20.36\% /\textbf{79.64\%}\\
    Film / \textbf{Ours} &24.77\% /\textbf{75.23\%}\\
    LDMVFI/ \textbf{Ours}&13.65\% /\textbf{86.35\%}\\
  \bottomrule
\end{tabular}
\end{minipage}
\end{table}

\subsection{User Study}
We further conducted a user study to investigate the performance of our method compared to all baselines from a human perspective.
Specifically, we collected $30$ pairs of images from~\emph{InterpBench}.
We used different approaches to generate videos with identical settings. 
During the study, we showed participants with the input image pair and two interpolation videos, one generated by our method and another randomly selected approach, in random order.
$139$ volunteers were invited to choose the method with better perceptual quality and realism. We report the results in Table~\ref{tab:user study}, which indicates that our method outperforms alternative approaches by a significant margin.
\subsection{Ablation Study}
\begin{figure}[!t]
  \centering
  \setlength{\belowcaptionskip}{-5pt}
  \includegraphics[width=0.99\linewidth]{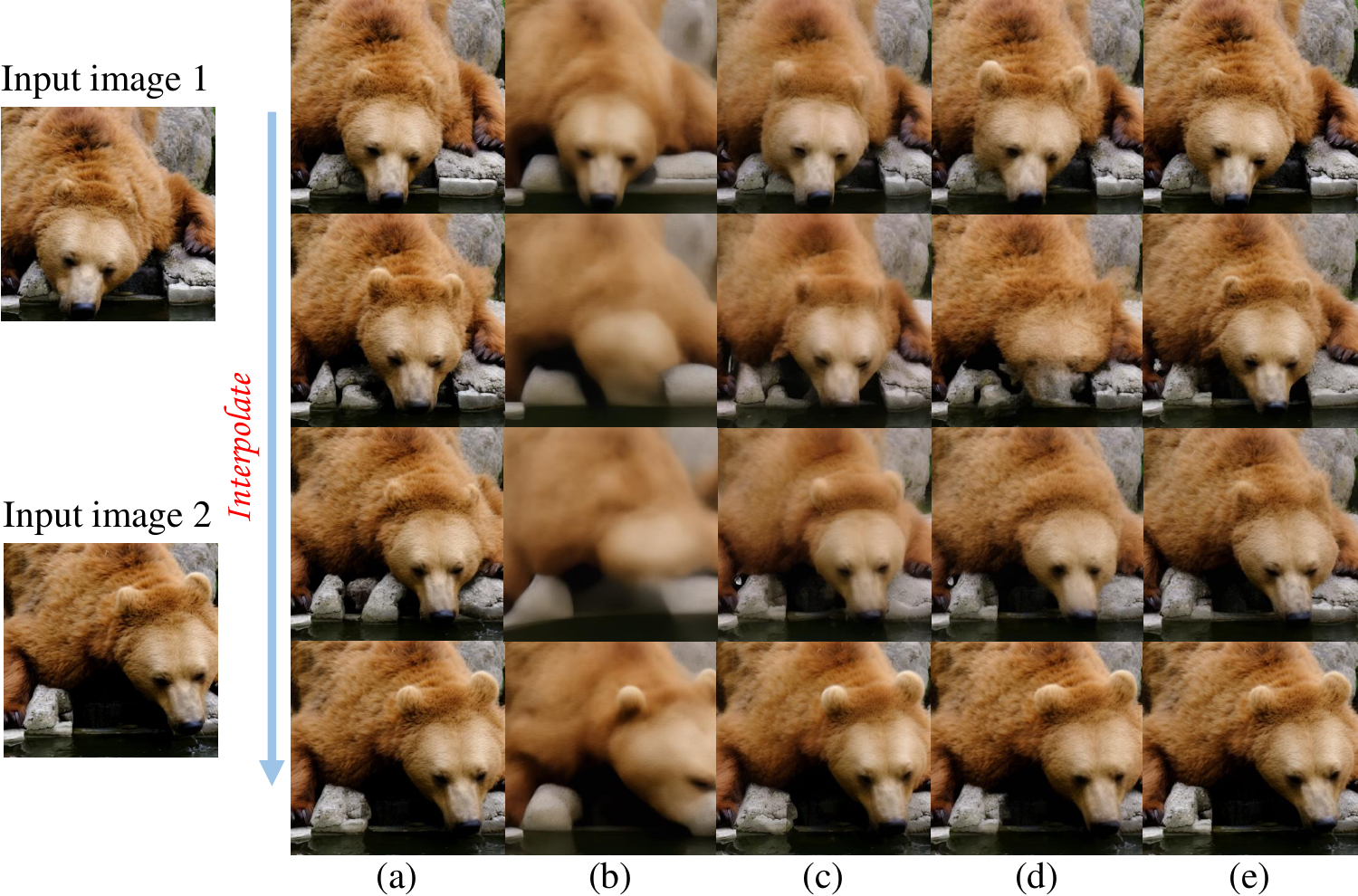}
  \caption{\textbf{Visual examples of the ablation study.} Each row shows the results of four intermediate images in different settings.}
  \label{fig:ablation}
\end{figure}
Each component of our system plays an important role in improving the generation quality. To justify our design choices, we conduct quantitative ablation studies, as presented in Table~\ref{tab:ablation}. Visual results of the ablation study are shown in Fig.~\ref{fig:ablation}.
In the ``w/o our estimator'' experiment, we employ ``off-the-shelf'' optical estimator RAFT~\cite{teed2020raft} as an alternative to obtain the flow. However, on one hand, directly acquiring the optical flow in the RGB space and using it in the latent space generates errors. On the other hand, it is difficult for general optical flow estimators to capture large motions. Therefore, as shown in Fig.~\ref{fig:ablation}(a), the rock on the left incorrectly moves with the bear since input image pairs are not derived with accurate correspondence by optical flow.
\begin{table}[htb]
\centering
  \caption{\textbf{Ablation Study on each component of our method.}}
  \label{tab:ablation}
  \resizebox{0.7\linewidth}{!}{
  \renewcommand\arraystretch{1.00}
  \begin{tabular}{lcccc}
    \toprule    Method&FID $\downarrow$ &LPIPS $\downarrow$ &WE $\downarrow$ &$\text{WE}_{mid}$ $\downarrow$ \\
    \midrule
    (a) w/o our estimator  &53.8510 &0.2334 & 0.2682 &0.5229\\
    (b) w/o two-level fusion &138.9329 &0.4286  &0.2693 &0.6510\\
    (c) w/o replace attention &69.4221 &0.2919 &0.2254 &0.4865\\
    (d) w/o lora   &76.7142 &0.2595  &0.2332  &0.4744       \\
    (e) Full model &\textbf{43.1798} &\textbf{0.2227} &\textbf{0.2069} &\textbf{0.3687}\\
  \bottomrule
\end{tabular}
}
\end{table}
In the ``w/o two-level fusion'', we directly fuse on $z_T$ space, and we can observe the results in Fig.~\ref{fig:ablation}(b) and Fig.~\ref{fig:fusion} are blurry. In the ``w/o replace attention'' and ``w/o lora'' experiments, the interpolation images are not consistent with the input two images. 
In the supplementary material, we discuss the effects of alternative fusion strategies and different total noisy timestamp $T$.
\section{Conclusion}
In this paper, we present a novel approach for image interpolation with large motion while ensuring the preservation of semantic consistency in the generated results.
By leveraging the prior knowledge of a pre-trained text-to-image diffusion model, we propose a natural optical flow estimator, a novel two-level fusion strategy, and a self-attention concatenation and replacement method to generate intermediate images. We conduct extensive experiments to verify the effectiveness of our method. We hope that our work can bring large motion interpolation into the sight of a broader community and motivate further research.

\noindent\textbf{Limitations.} 
Our method leverages the prior of the pre-trained diffusion model, but meanwhile inherits its limitations. Since we employ the diffusion model at low resolution latent space, it may cause texture sticking and be difficult to capture slight motion. We plan to explore more effective solutions in future work.
% , it is difficult for our approach to be spatially aware. 
% Now we have reached the maximum length of an ECCV \ECCVyear{} submission (excluding references).
% References should start immediately after the main text, but can continue past p.\ 14 if needed.
\clearpage  % TODO REVIEW/FINAL: This \clearpage needs to be removed from both review and camera-ready versions.

\bibliographystyle{splncs04}
\bibliography{main}
\end{document}